\documentclass[conference]{IEEEtran}
\ifCLASSINFOpdf
\else
\fi
\usepackage{amsmath}

\usepackage{caption} 
\usepackage{booktabs}
\usepackage{enumitem} 
\usepackage{wrapfig} 
\usepackage{tcolorbox} 
\usepackage{blkarray}
\usepackage{multirow}
\usepackage{soul}
\usepackage{tabularx}
\usepackage{array}
\usepackage{colortbl}
\tcbuselibrary{skins}
\usepackage{soul}
\usepackage{algorithm}
\usepackage{algcompatible}
\usepackage{algpseudocode}
\usepackage{amsfonts}

\usepackage[T1]{fontenc}
\usepackage[font=small,labelfont=bf,tableposition=top]{caption}

\usepackage{capt-of}

\usepackage{accents}

\usepackage{algcompatible}
\usepackage{algpseudocode}
\usepackage{pifont}

\PassOptionsToPackage{hyphens}{url}\usepackage{hyperref}

\newcolumntype{Y}{>{\raggedleft\arraybackslash}X}
\tcbset{tab1/.style={fonttitle=\bfseries\large,fontupper=\normalsize\sffamily,
colback=yellow!10!white,colframe=red!75!black,colbacktitle=Salmon!40!white,
coltitle=black,center title,freelance,frame code={
\foreach \n in {north east,north west,south east,south west}
{\path [fill=red!75!black] (interior.\n) circle (3mm); };},}}
\tcbset{tab2/.style={enhanced,fonttitle=\bfseries,fontupper=\normalsize\sffamily,
colback=yellow!10!white,colframe=red!50!black,colbacktitle=Salmon!40!white,
coltitle=black,center title}}
\newcolumntype{x}{>{\centering\arraybackslash}X}
\newcolumntype{s}{>{\centering\arraybackslash}>{\hsize=.25\hsize}X}

\newcommand{\ds}{\displaystyle}
\newcommand{\df}{\displaystyle\frac}

\newcommand{\+}[1]{\ensuremath{{\boldsymbol #1}}} 
\newcommand{\quotes}[1]{``#1''}
\newcommand{\R}{\mathbb{R}}

\newcommand{\explain}[2]{\underbrace{#1}_{\text{#2}}}

\newcommand{\pluseq}{\mathrel{+}=}

\makeatletter
\newcommand*\bigcdot{\mathpalette\bigcdot@{.5}}
\newcommand*\bigcdot@[2]{\mathbin{\vcenter{\hbox{\scalebox{#2}{$\m@th#1\bullet$}}}}}
\makeatother

\begin{document}
%
\title{Projecting ``better than randomly": How to reduce the dimensionality of very large datasets in a way that outperforms random projections}

\author{\IEEEauthorblockN{Michael Wojnowicz, Di Zhang, Glenn Chisholm, Xuan Zhao, Matt Wolff}
\IEEEauthorblockA{Department of Research and Intelligence \\
Cylance, Inc. \\
Irvine, California 92612 \\
\{mwojnowicz, dzhang, gchisholm, xzhao, mwolff\}@cylance.com}

}

\maketitle

\begin{abstract}
For very large datasets, random projections (RP) have become the tool of choice for dimensionality reduction. This is due to the computational complexity of principal component analysis.    However, the recent development of randomized principal component analysis (RPCA) has opened up the possibility of obtaining approximate principal components on very large datasets.  In this paper, we compare the performance of RPCA and RP in dimensionality reduction for supervised learning.   In Experiment 1, study a malware classification task on a dataset with over 10 million samples, almost 100,000 features, and over 25 billion non-zero values, with the goal of reducing the dimensionality to a compressed representation of 5,000 features.  In order to apply RPCA to this dataset, we develop a new algorithm called large sample RPCA (LS-RPCA), which extends the RPCA algorithm to work on datasets with arbitrarily many samples.  We find that classification performance is much higher when using LS-RPCA for dimensionality reduction than when using random projections.  In particular, across a range of target dimensionalities, we find that using LS-RPCA reduces classification error by between 37\% and 54\%.  Experiment 2 generalizes the phenomenon to multiple datasets, feature representations, and classifiers.   These findings have implications for a large number of research projects in which random projections were used as a preprocessing step for dimensionality reduction.  As long as accuracy is at a premium and the target dimensionality is sufficiently less than the numeric rank of the dataset, randomized PCA may be a superior choice.  Moreover, if the dataset has a large number of samples, then LS-RPCA will provide a method for obtaining the approximate principal components. 
\end{abstract}


\section{Introduction}

We consider an increasingly typical data analytic situation where one wants to perform supervised learning on a very large dataset: one with both many samples and many features, and which may be too large to fit into memory.  Because computationally demanding classifiers, such as neural networks, can struggle with high-dimensional feature spaces, it is common to first preprocess the data with a dimensionality reduction technique.   


Although the classical dimensionality reduction technique is principal components analysis, it is computationally intensive; therefore, for large datasets, the use of random projections for dimensionality reduction has become nearly ubiquitous.  Random projections are a computationally cheap, and surprisingly effective, way to reduce dimensionality without a large loss of information.  Random projections ``sketch" a large data matrix by taking a small number of (randomly weighted) linear combinations of the rows or columns of that matrix.   The method allows one to solve data analytic problems in lower-dimensional spaces that tend, counter-intuitively, to provide  (provably) good approximations to the solutions in the original space~\cite{mahoney}.  Thus, a relatively common workflow for performing supervised learning on large datasets is as follows: (1) perform a random projection on the dataset, (2) feed the reduced features into a (possibly expensive) classifier.  For instance, Dahl. et al built an effective malware classifier on 2.6 million samples and 179,000 features by first performing a random projection, and then feeding the projected features into a neural network~\cite{dahl}. 

However, the widespread use of random projections on large datasets may have suboptimal consequences.  Principal component analysis is the optimal linear dimensionality reduction technique (from a number of perspectives, such as preserving variance explained).  Thus, principal components analysis should provide gold-standard projection matrices\footnote{Technically speaking, principal component reductions are not projections; they are rotations of the dataset (after which low-variability dimensions are discarded).  By a similar argument, random ``projections" are also not projections.  However, in this paper, we follow the convention of the literature on random projections and refer to both as ``projections," in the loose sense that they reduce dimensionality by combining the original features.}, and random projections may or may not match their efficacy. Indeed, at an empirical level, a number of papers have directly compared the performance of random projections (RP) and principal component analysis (PCA) as dimensionality-reducing preprocessing steps for supervised learning algorithms, and have found that principal component analysis provides better dimensionality reduction for downstream classifiers.  For example, Fradkin et al.~\cite{fradkin}  found that, on a variety of machine learning datasets, PCA outperforms RP as a preprocessing step for nearest neighbor classifiers, support vector machines, and decision trees.  Deegalla et al.~\cite{deegalla} confirmed this result for nearest neighbor classifiers.    This research suggests that the two dimensionality reduction techniques present a trade-off between computational time and accuracy, with PCA being favored when high accuracy is at a premium.

But what if the datasets are large? The aforementioned studies comparing PCA to random projections were made on relatively small datasets (for instance, no dataset analyzed by Deegalla et al. included more than 9,000 samples or 7,200 predictors).  For datasets that are large  (e.g. Dahl et al.'s malware dataset, with tens of million of samples and hundreds of thousands of predictors), the computational complexity of ordinary, deterministic PCA can be too prohibitive.  In particular, the classical method for obtaining the principal components of an $N \times P$ dataset (where the rows are samples and columns are predictors) would require $O(NP\;min\{N,P\})$ computations~\cite{golub}.  For datasets whose numerical rank is $K < min\{N,P\}$, the run time can be reduced to approximately $O(NPK)$ using Krylov subspaces, but these methods can be numerically unstable, and would require $K$ passes over the dataset, which can be prohibitive if the dataset is stored out-of-core and if $K$ is large~\cite{halko_big}.    In contrast, a random projection to a target dimensionality of $K$ variables requires a simple matrix multiplication: \emph{at most} $O(NPK)$ computations in a single pass through the dataset.\footnote{Certain choices for the random projection matrix can reduce this even further to $O(NP\log{K})$ computations. Taking advantage of sparsity can reduce the computational complexity as well.~\cite{mahoney}}   Therefore, by default, the dimensionality reduction method of choice, for large datasets, has tended to be random projections. 

 
However, thanks to the relatively recently developed framework of randomized principal components analysis (RPCA)~\cite{halko_pca}, it is now possible to obtain \emph{approximate} principal components for very large datasets, even those which do not fit into memory.   RPCA finds approximate principal components that, because of concentration of measure results~\cite{halko_big}, are provably close to the true principal components.     As with random projections, RPCA is a stochastic algorithm, but RPCA employs stochasticity in a more focused way.   Whereas for random projections, the projection matrix is formed completely randomly and independently of the data, for RPCA, randomness is used specifically to help approximate the column space of the data matrix; then, that information and the original data itself are combined to determine the approximate principal components.  

While the previously discussed research (\cite{deegalla},\cite{fradkin}) has found that,  for small datasets, \emph{deterministic} PCA can provide better projections than random projections, we are unaware of work that has posed the analogous question for large datasets, comparing the effectiveness of \emph{randomized} PCA to random projections.  As we discuss in Section~\ref{compare}, strong arguments can be made both ways about whether and when RPCA should provide better projections than random projections.  But, in brief, the randomization of RPCA may cause PCA to lose its relative advantage over RP, and the advantage of PCA over RP may not be strong to begin with.  Thus, we pose the following question:


  
  
\begin{quote} 
\underline{Question of Interest \#1}: \emph{Can randomized principal component analysis (RPCA) reduce the dimensionality of a large-scale dataset  in a way that outperforms random projections with respect to downstream classification?}  
\end{quote} 

That is, can we project large datasets better than randomly? 

As we will see in Section~\ref{introducingLSRPCA}, when datasets are sufficiently large, even the \emph{randomized} PCA algorithm is not necessarily straightforward to apply.   In essence, the standard algorithm for RPCA~\cite{halko_pca} requires an in-core QR decomposition on a dense matrix with $N$ samples and $K$ reduced features.  This can easily become infeasible; for instance, Julia's native QR algorithm implemented on an Amazon EC2 r3.4 instance with 120 GB of RAM produces out-of-memory errors on a  simulated Float32 matrix with 1.5 million samples and a target dimensionality of K=$5000$.  As a result, for these parameter values, it would not be possible to run the standard algorithm for RPCA.   Researchers who find themselves analyzing high-dimensional datasets that also have many samples may therefore give up, turning automatically to random projections for dimensionality reduction, even though this choice may produce suboptimal low-dimensional representations relative to RPCA. Thus, we pose a second question as well:    




\begin{quote} 
\underline{Question of Interest \#2}: \emph{Can we develop a \quotes{large-sample} variant of RPCA, which gracefully handles high-dimensional datasets with many samples?}  
\end{quote}

 

\subsection{Dimensionality Reduction Strategies}

\subsubsection{Notation}
We represent our dataset as a matrix $\+X \in \R^{N \times P}$, where $N$ is the number of samples and $P$ is the number of predictors.     We use $\+x_i$ to refer to the $i$th column of $\+X$ and $\+x_i^T$ to refer to the $i$th row of $\+X^T$.   We use $K$ to refer to the \quotes{target dimensionality} (i.e., we'd like to  reduce the dimensionality of $\+X$ to  $K<P$ predictors).  Below we describe a number of methods for doing so.  The quantity $\overline{K}$ refers to an ``over-sampling" dimensionality for the RPCA algorithm; it is a number that is typically slightly bigger than $K$.  
 
%

\subsubsection{PCA} \label{PCA}
It is well known that principal component analysis yields the \quotes{optimal} linear method for reducing the dimensionality of a dataset in terms of preserving variance explained.  In particular, we can obtain the principal components through a singular value decomposition. Letting $\+X = \+U\+\Sigma\+V^T$, the \emph{principal directions} are given by $\+V$, and the \emph{principal components} or \emph{principal component scores} are given by $\+X\+V \in \R^{N \times P}$.

To obtain what we are calling the \emph{principal components projection}, that is the low-dimensional representation $\+X^{\text{proj}}\in \R^{N \times K}$ analogous to the low-dimensional representation of random projections,   we first approximate $\+X$ with a rank $K$ matrix, $\+X_K$, obtained by performing a singular value decomposition (ordered such that that the singular values are non-decreasing) and then truncating to the $K$ dominant singular values $\{\sigma_1, \hdots \sigma_K \}$ and corresponding $K$ dominant left and right singular vectors:
\begin{equation}
  \label{rank_k}
  \+X_K := \+U_K\+\Sigma_K\+V_K^T= \ds\sum_{i=1}^K \sigma_i \+u_i \+v_i^T 
  \end{equation}
  
This rank-k matrix $\+X_k$, obtained from the truncated SVD, is the optimally best rank-k approximation to the original complete dataset, in the sense that: 
\[ || \+X - \+X_K || = \min_{rank(\+A)=K} \; ||\+X- \+A||=  \sigma_{K+1} \]
where $||\cdot||$ can be any unitarily invariant norm, such as the Frobenius or the L2 norm.  From this, we can obtain the principal components projection via \[\+X^{\text{proj}} = \+X\+V_K = \+U_K\+\Sigma_K\]

%

\subsubsection{Randomized PCA (RPCA)} \label{RPCA_section}

\begin{algorithm}
\caption{A Baseline RPCA Algorithm~\cite{halko_pca}}
\label{RPCA}
\begin{algorithmic}
\State {\bf Data} A dataset $\+X \in \R^{N \times P}$ \\
Target dimensionality $K$ \\
Oversampled target dimensionality $\overline{K}$ 
\State {\bf Result} Projection matrix $\+V \in \R^{P  \times K}$  \\

\State 1. Form random projection matrix $\explain{\+\Omega}{$P \times \overline{K}$}$
\State 2. Compute  $\explain{\+Y}{$N \times \overline{K}$} := \explain{\+X}{$N \times P$}  \hspace{.05in}  \explain{\+\Omega}{$P \times \overline{K}$} \; $
\State 3. Do QR decomposition: $\explain{\+Q}{$N \times \overline{K}$}  \hspace{.05in}  \explain{\+R}{$\overline{K} \times \overline{K}$} = \explain{\+Y}{$N \times \overline{K}$}$
\State 4. Compute $\explain{\+B}{$\overline{K} \times P$} := \explain{\+Q^T}{$\overline{K} \times N$} \hspace{.05in} \explain{\+X}{$N \times P$} $
\State 5. Do SVD of $\+B$: $\explain{\+B}{$\overline{K} \times P$} = \explain{\widetilde{\+U}}{$\overline{K} \times \overline{K}$} \hspace{.05in} \explain{\+\Sigma}{$\overline{K} \times \overline{K}$}  \hspace{.05in} \explain{\+V^T}{$\overline{K} \times P$} $
\State 6. Now we can get approx. SVD of $\+X$ by multiplying $\+Q\widetilde{\+U}$:
\begin{align*}
\+X & \approx  \explain{\+Q}{$N \times \overline{K}$} \hspace{.05in}    \explain{\+Q^T}{$\overline{K} \times N$} \hspace{.05in}   \explain{\+X}{$N \times P$} \\
&=   \explain{\+Q}{$N \times \overline{K}$} \hspace{.05in}    \bigg( \explain{\widetilde{\+U}}{$\overline{K} \times \overline{K}$} \hspace{.05in}    \explain{\+\Sigma}{$\overline{K} \times \overline{K}$} \hspace{.05in}   \explain{\+V^T}{$\overline{K} \times P$} \bigg) \\
&:=  \explain{\+U}{$N \times \overline{K}$} \hspace{.05in}    \explain{\+\Sigma}{$\overline{K} \times \overline{K}$} \hspace{.05in}   \explain{\+V^T}{$\overline{K} \times P$} \\
\end{align*}
\State 7. Project the dataset
\[\+X^{\text{proj}} = \+X \+V_K \]
where $\+V_K$ is $\+V$ truncated column-wise to $K$ leading dimensions
\end{algorithmic}
\end{algorithm}

Randomized PCA provides an approximation to PCA, which can be computationally infeasible on large datasets. The method~\cite{halko_big},~\cite{halko_pca} yields an approximation to the rank-$K$ truncated SVD  $\+X_k \in \R^{N \times K}$ provided in Equation~\ref{rank_k} and from which the principal components and principal component projection can be easily derived.  

The main idea is to approximate $\+X$ with the matrix $\+Q\+Q^T\+X$, where $\+Q  \in \R^{N \times \overline{K}}$ is a matrix with $\overline{K}$ orthonormal columns that approximates the column space or range of $\+X$, and where $\overline{K}$ is slightly bigger than the target dimensionality, $K$.   Accomplishing this approximation involves randomly projecting the data matrix $\+X$ to $\overline{K}<P$ dimensions, and then forming a QR decomposition on the projected dataset.    With this approximation $\+X \approx \+Q\+Q^T\+X$ in hand, one can take the SVD of an $\overline{K} \times P$ matrix to obtain an approximate SVD for the much larger $N \times P$ matrix.   In Algorithm~\ref{RPCA}, we provide the standard, baseline algorithm from~\cite{halko_big},~\cite{halko_pca}.    

Note that there are various other instantiations of this algorithm that may be appropriate depending on the context (e.g. other kinds of matrices can be used in Step 1; see~\cite{halko_big} for a discussion).  In particular, we do not include \quotes{power iterations} in the standard algorithm.  Power iterations involve premultiplying the dataset, $\+X$ with the premultiplier $\+X \+X^T$  numerous times before random projection; that is, power iterations involve setting $\+Y=(\+X\+X^T)^q \+X\+\Omega$ in Step 2  of Algorithm~\ref{RPCA} for some natural number $q$.   The purpose of these power iterations is to accelerate singular value decay for datasets with a relatively flat spectrum (i.e, datasets where many dimensions would be needed to capture a sufficiently large percentage of variation in the samples).   In such a setting, power iterations would allow the randomized PCA to provide better approximations with better error guarantees. However, for datasets with many (e.g. 10 million) samples, performing these power iterations can be computationally infeasible.  Thus, as we consider RPCA on datasets with very large datasets which may have many millions of samples, we drop this option from the standard algorithm and from our approach.  


\subsubsection{Random Projections}
Random projections have been developed as extensions of the Johnson-Lindenstrauss (J-L) lemma, which implies that, with high probability, the low-dimensional representations of samples will preserve the original pairwise distances between them (within a margin of error that depends upon the number of reduced dimensions; see Equation~\ref{rp_error}.)  By a random projection, we mean 
\[ \+X^{\text{proj}} = \df{1}{\sqrt{K}} \+X \+\Omega \]
where $\+\Omega \in \R^{P \times K}$ is a matrix of random numbers.  There are many possible methods for constructing the random matrix. For this paper, we focus on most prevalent procedure, which is to make $\+\Omega$ Gaussian (where each element is an i.i.d draw from a standard normal distribution, i.e. $\Omega_{i,j} \sim N(0,1)$.)  In practice, for large scale applications, similar results can be obtained with reduced storage and computation costs using more contemporary methods such as a \emph{very sparse random projection}~\cite{li}.

\section{A priori comparisons} \label{compare}
\subsection{Randomized PCA vs. random projections: Which should we expect to perform better?} \label{compare}
Because PCA is the optimal linear dimensionality reduction technique, it seems plausible, on the face of it, that a direct approximation to PCA would outperform any other method, such as random projections.  However, a deeper look reveals that it is by no means obvious that randomized PCA would outperform random projections at dimensionality reduction.  There are a number of reasons for uncertainty:  

\begin{enumerate}[leftmargin=*]

\item \emph{RPCA approximates PCA}.  Although a truncated PCA provides an optimal linear projection, RPCA is a randomized \emph{approximation} to that.   Given a dataset $\+X$ with $N$ samples and $P$ predictors, and applying Algorithm~\ref{RPCA} with $\overline{K}=2K$, the RPCA method produces an approximating matrix $\+X_K = \+U_K \+\Sigma_K \+V_K$, from which the approximate principal components can be derived, as shown in Section~\ref{PCA}.  According to~\cite{halko_big}, the approximating matrix $\+X_K$ has error guarantee

\begin{equation}
\label{rpca_error}
\mathbb{E} || \+X- \+X_K || \leq \bigg(2 + 4 \sqrt{\df{2\; \text{min}\{N,P\}}{K-1}} \bigg) \sigma_{K+1}
\end{equation}
where $\sigma_{K+1}$ is the $(K+1)st$ largest singular value of $\+A$.  This upper bound is larger than the error of an approximating matrix formed by deterministic PCA, where $ \mathbb{E} || \+X - \+X_K || = \sigma_{K+1}$.  How do we know that the approximating quality of RPCA doesn't destroy the advantages of PCA, which makes it preferable to RP in the first place? 

\item \emph{RP can approximate PCA.}  The effectiveness of random projections is often counter-intuitive, since random projection matrices are formed without any reference to the underlying data.    Figure~\ref{MagixOfRandomProjections} shows a 2-dimensional dataset with a strong correlation between features, where the samples are colored roughly according to their ordering along the dominant principal component.  People's intuition may hold that random projection, being random, must project points in a uniformly distributed manner between the optimal projection to the first principal component (where colorings would be largely preserved), and the disastrous projection to the second principal component (where the colors would be almost randomly intermingled).   As it turns out, this is is wrong.  As the plot shows, the random projection, although random, provides almost identical low-dimensional information as does the linearly optimal projection to the dominant principal component!

\item \emph{Difficulty applying theoretical error guarantees for RPCA and RP.}  Good behavior for both RPCA and RP is guaranteed by error bounds. Equation~\ref{rpca_error} provides relevant error bound for RPCA.   For comparison, consider a RP, using a Gaussian random projection matrix, to $K$ dimensions.  Then by ~\cite{arriaga}, the J-L lemma yields the following \emph{preservation-of-distance} guarantee: for any $\epsilon>0$ and any samples $\+x_i^T,\+x_j^T \in \R^{P}$, the projected samples $\+x_i^{T\prime},\+x_j^{T\prime} \in \R^K$ satisfy:
\begin{equation}
\label{rp_error}
\mathbb{P}\bigg( ||\+x_i^{T\prime} - \+x_j^{T\prime} ||^2  \in (1 \pm \epsilon) ||\+x_i^T-\+x_j^T||^2 \bigg) \leq 1- 2e^{-(\epsilon^2-\epsilon^3)K/4}
\end{equation}
It is not completely clear how to apply these error bounds to compare performance of RP against RPCA.  For example, random projections and PCA subserve different goals.  By equation~\ref{rp_error}, random projections limit the extent to which projections distort the distances of \emph{all} points.  On the other hand, RPCA, like PCA, attempts to minimizes the extent to which projections distort the \emph{average} point.  (See, e.g., Equation~\ref{rpca_error}.)   Moreover, the bounds described by the theorems may of course differ in how loose they are relative to particular applications.

 \begin{figure}
\centering
\includegraphics[height=3.2in]{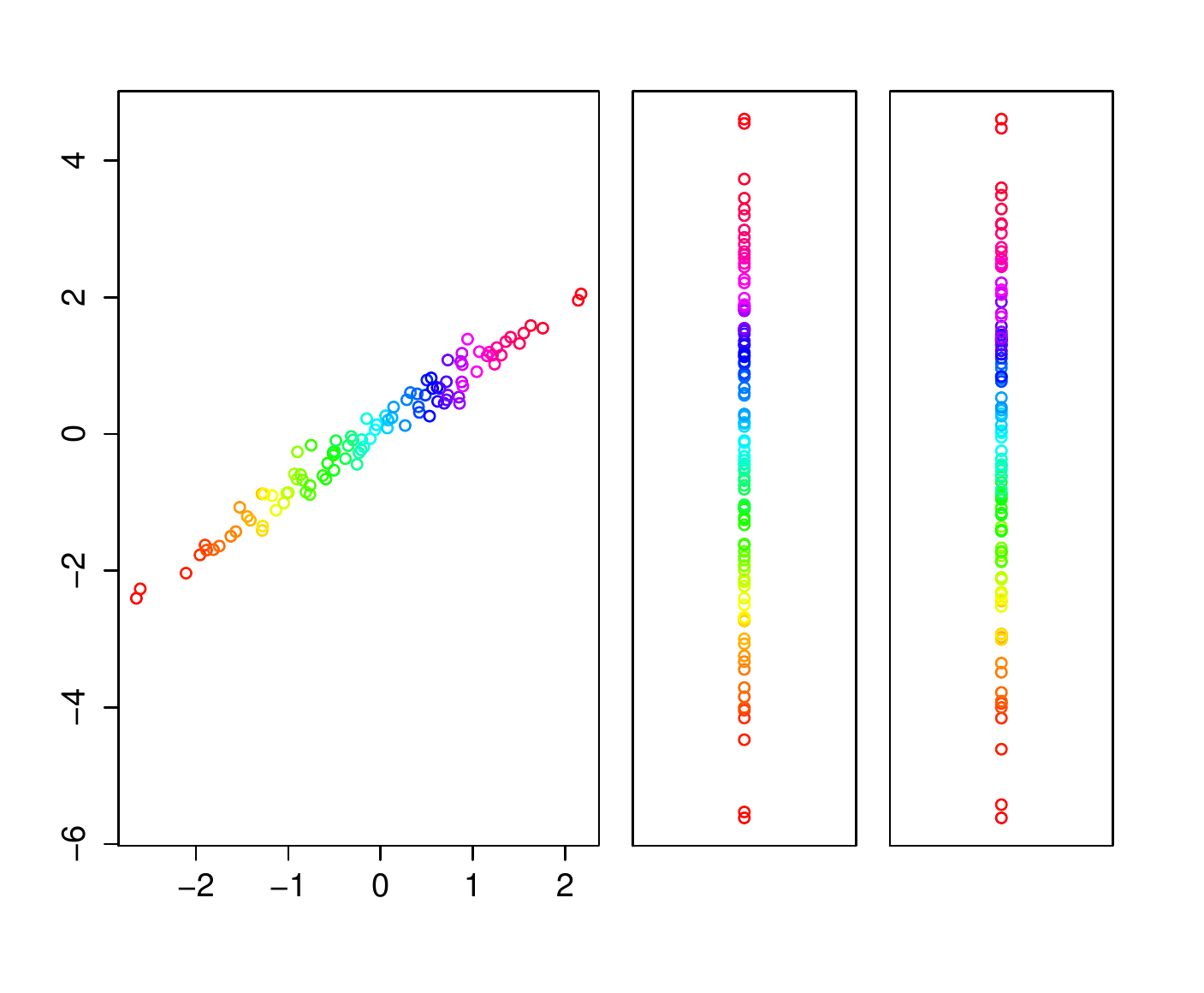}
\caption{ \emph{The magic of random projections.} The left panel shows a 2-dimensional point cloud with strong correlation between features.  For the two panels on the right, one shows a representative 1-dimensional representation after projecting to the dominant principal component, and the other shows a representative 1-dimensional representation after a random projection. Which is which?}
\label{MagixOfRandomProjections}
\end{figure}

\item \emph{Random projections can introduce useful distortions.}   There is a concentration of measure theorem for random projections which states that the randomly projected data, at least when projected with Gaussian matrices, looks like scale-mixtures of spherical Gaussians~\cite{dasgupta}.    In contrast, the principal component representation can show strange nonlinear dependencies (see Figure 6 of~\cite{dasgupta}, which shows both kinds of projections for various ML datasets).  The limiting effects of random projections might be considered distortions; alternatively, the well-behavedness of nearly spherical Gaussian data can actually be beneficial for downstream applications.  (For example in a paper on clustering high-dimensional data, random projections (at least when ensembled) led to better low-dimensional clusters than principal components \cite{fern}.)  It is unclear which perspective is dominant in the context of prediction, with the answer likely depending upon the particular dataset and the classifier. 

\end{enumerate} 


\section{How to extend RPCA to datasets with large $N$} \label{introducingLSRPCA}
We contribute a new algorithm, which we call large sample Randomized Principal Components Analysis (LS-RPCA).  The LS-RPCA algorithm extends the randomized PCA framework to data sets with very many samples, without having to subsample or discard data. 

The standard RPCA algorithm, as stated in~\cite{halko_pca} does not  scale to data sets with many samples (large $N$).   To see why, note that the standard RPCA algorithm requires an in-core qr decomposition on an $N \times K$ matrix~\cite{halko_pca}.  Typically $K$ is relatively small, but if $N$ is large, then the QR algorithm would not be possible to do in-core.  

 
 We resolve the large-sample bottleneck by adjusting the randomized PCA algorithm to gracefully accomodate an out-of-core QR algorithm~\cite{gunter},~\cite{buttari} directly into the randomized PCA algorithm. The algorithm we develop computes RPCA while simultaneously (and not sequentially) performing an out-of-core QR algorithm.  In this way, our algorithm minimizes storage costs and maximizes pass-efficiency, requiring only a single pass through the dataset.   (This is important because when datasets are so large that they do not fit into fast memory, computation time is typically dominated by memory access rather than floating-point operations~\cite{halko_big}) 
 
%
%
 Whereas the bottleneck for RPCA is the ability to operate upon an $N \times K$ matrix in core, the bottleneck for LS-RPCA is to operate upon an $P \times K$ matrix in core.\footnote{The LS-RPCA algorithm, like the standard RPCA algorithm, still requires an in-core SVD on a $P \times K$ matrix.}     In fact, LS-RPCA \emph{removes any restriction on $N$},  so long as one is willing to absorb the linear run-time dependence on $N$.
 
  
 


\subsection{Out-of-core QR decompositions}

Out-of-core QR decompositions are built off the method of QR decomposition via Householder reflections, which we briefly review here. 

\subsubsection{QR decompositions via Householder reflections} \label{QRHouseholder}

Let $\+A$ be an $M \times N$ matrix.  A \emph{full QR decomposition} is $\+A=\+Q\+R$ where $\+Q$ is an $M \times M$ orthogonal matrix and $\+R$ is an $M \times N$ upper triangular matrix. If $M>N$, the last $M-N$ rows of $\+R$ will be zero, so we can also form a \emph{reduced (or thin) QR decomposition},  $\+A=\+Q\+R$ where $\+Q$ is an $M \times N$  matrix with orthonormal columns and $\+R$ is an $N \times N$ upper triangular matrix.   

Here we briefly review the construction of the QR decomposition using Householder reflections, as this method provides the basis for the out-of-core (or \quotes{tiled}) QR decomposition.   Following~\cite{golub}, we compute  $\+R=\+Q\+A$, where $\+R$ is upper triangular and $\+Q$ is orthogonal.   We will construct $\+R$ iteratively:  after the $i$th iteration, we will have constructed a matrix $\+R_i$ which is upper triangular for columns $1,\hdots,i$.   
Now the Householder reflection theorem tells us that if $\+x$ and $\+y$ are two vectors with the same norm, then there exists an orthogonal, symmetric matrix $\+Q$ such that $\+y=\+Q\+x$ (and it tells us how to construct such a matrix).   We apply this theorem to determine an orthogonal matrix,  $\+Q_1$, such that $\+Q_1 \+a_1 = (\gamma_1,0,0,\hdots,0)^T$, where $\gamma_1$ is equivalent to the norm of $\+a_1$.  If we apply $\+Q_1$ to the entire matrix $\+A$, we obtain $\+R_1 = \+Q_1 \+A$, where
\[\+R_1 = \begin{blockarray}{c|ccc}
    \begin{block}{(c|ccc@{\hspace*{5pt}})}
     \gamma_1 & \times & \times & \times \\
     \cline{1-4} 
     0 & \BAmulticolumn{3}{c|}{\multirow{3}{*}{$\+A_1$}}\\
    0 & &&&&&&&\\
    0 & &&&&&&&\\
    \end{block}
  \end{blockarray} 
\]

Now we perform the same process as before on this smaller dimensional matrix $\+A_1 \in \R^{(M-1) \times (N-1)}$.  The Householder reflection theorem guarantees the existence of an orthogonal matrix $\tilde{\+Q}_2 \in \R^{(M-1) \times (N-1)}$ such that $\tilde{\+Q}_2 \tilde{\+a}_1 = (\gamma_2,0,0,\hdots,0)^T$ $\in \R^{M-1}$.   We pad this to create a matrix $\+Q_2 = diag(\+I_1,\tilde{\+Q}_2)  \in \R^{M \times N}$, and multiply it by our in-progress triangular matrix, $\+R_1$, to get the next iteration of our triangular matrix $\+R_2$, which will have the first two columns set:

\small
\begin{align*}
\+R_2 = \+Q_2 \+R_1 = \begin{blockarray}{c|ccc}
    \begin{block}{(c|ccc@{\hspace*{5pt}})}
     \gamma_1 &  \times  & \hdots &  \times  \\
     \cline{1-4} 
     0 & \BAmulticolumn{3}{c|}{\multirow{3}{*}{$\tilde{\+Q}_2 
     \+A_1$}}\\
    \vdots & &&&&&&&\\
    0 & &&&&&&&\\
    \end{block}
  \end{blockarray} 
   \;  = \; \begin{blockarray}{cc|cc}
    \begin{block}{(cc|cc@{\hspace*{5pt}})}
     \gamma_1 &  \times  &  \times  & \times  \\
        0 & \gamma_2 &  \times &  \times  \\
     \cline{1-4} 
     0 & 0& \BAmulticolumn{2}{c|}{\multirow{2}{*}{$\+A_2$}}\\
    \vdots & \vdots &&&&&&&\\
    0 & 0 &&&&&&&\\
    \end{block}
  \end{blockarray} \\
\end{align*}
\normalsize

We can iterate this $t$ times, where $t=min\{m-1,n\}$, after which point we have $\+R = \+Q_t \+Q_{t-1} \cdots \+Q_2 \+Q_1 \+A$. Note that the $\+Q_t$'s are orthogonal, based on how we constructed them from the $\tilde{\+Q}_t$'s, which are orthogonal by the Householder Reflection Thm.   Thus, by orthogonality, we have $\+Q_t^T \+Q_t = \+I$. So, taking the inverses, we obtain $ \+A= \+Q_1^T \+Q_2^T \cdots \+Q_{t-1}^T \+Q_t^T \+R  := \+Q\+R$, where it is very easy to check (via condition $\+Q^T\+Q=I$) that $\+Q:=\+Q_1^T \+Q_2^T \cdots \+Q_{t-1}^T \+Q_t^T $ is an orthogonal matrix. 

\subsubsection{Tiled QR Decompositions}

Tiled QR decompositions (~\cite{buttari},~\cite{gunter})  are designed for \quotes{out-of-core} (OOC) performance, because a matrix $\+A$ may not fit into RAM.  Tiled QR Decompositions work by bringing in only certain pieces of the original matrix $\+A$ into memory at a time, operating upon them, and proceeding until a QR decomposition has been performed on the full matrix.   In algorithm~\ref{CHalgorithm}, we present the main idea of these algorithms,  abstracting out  specific details like Compact WYQ representations and specific names of LAPACK function calls.   

 Essentially, we imagine $\+A \in \R^{M \times N}$ is divided up into a set of blocks or \quotes{tiles}; $\+A=(\+A_{ij})$, where $\+A_{ij}$ denotes the (i,j)th block of $\+A$, and where the $(i,j)$th block has size $M_{ij} \times N_{ij}$.  Furthermore, we assume that $M_{ii} > N_{ii}$ for each $i$ (that is, we assume that we have partitioned up the matrix in such a way that the diagonal blocks are tall-and-skinny).   The idea of how to perform a tiled QR follows naturally from QR-via-Householder-reflection procedure of Section~\ref{QRHouseholder}.   As before, we incrementally convert the $\+A$ matrix into an $\+R$ matrix by constructing a sequence of orthogonal matrices that operate only on a particular part of the matrix.    We zero out the appropriate blocks column-wise from left to right, analogously to the original QR algorithm, with the extra twist that we must ``couple" two blocks together (e.g., in the first column, we couple the (1,1) blocks with the blocks beneath it).  The purpose of the coupling is that since the upper block, which is always an $(i,i)$th block, is tall-and-skinny, the QR step will zero out the lower blocks beneath it.  

In particular, we begin with tile $\+A_{1,1}$.  By Householder's Theorem (and the padding trick of the previous section), we can construct an orthogonal matrix $\+Q_{11} \in \R^{M \times N}$ such that applying it to the matrix $\+A$ converts $\+A_{11}$ into an upper triangular matrix (so we now call it $\+R_{11}$).\footnote{Note that for this to work, we must construct the blocks so that each block is \quotes{tall-and-skinny}; i.e. each block has more rows than columns.}     We multiply through, noting that $\+Q_{11}$ only transforms tiles in the first row of tiles $\{ \+A_{1,j} \}_{j =1,\hdots,n}$.  Note the practical relevance for an out-of-core algorithm: we have so far only needed to transform tiles in the first row of tiles, and we can transform them one tile at a time.  

Now we proceed downward to consider tile $\+A_{21}$.   To obtain an upper triangular matrix in the end, we need to zero out this block.  To do so, we vertically concatenate $\+R_{11} $ and $\+A_{21}$ into a single matrix $\+B_{11} : = [\+R_{11}; \+A_{21}] \in \R^{(M_{11}+M_{21}) \times (N_{11}+N_{21})}$, and Householder's Theorem provides an orthogonal matrix $\+Q_{21}$ which converts the block $\+B_{11}$ into an upper triangular matrix (and therefore zeros out $\+A_{21}$).    Constructing $\+Q_{21}$ required seeing only $\+R_{11}$ and $\+A_{21}$, and the remaining matrix multiplication can be done tile-wise 
on tiles  $\{ \+A_{i,j} \}_{i=1,2; j =1,\hdots,n}$. We now iterate down the first column of tiles of A, such that we vertically concatenate $\+R_{11}$ and $\+A_{i1}$ for each $i= 2,\hdots,N$, construct the corresponding orthogonal transformation matrices $\+Q_{i1}$, and transform the relevant two rows of the $A$ matrix through matrix multiplications.  

At this point, the first column of tiles has been transformed into an upper triangular shape, and we move to the second column.  We proceed in this way through all columns, noting that for the $K$th column, we can ignore all tiles in rows $i<k$ (again, paralleling Section~\ref{QRHouseholder}), as these tiles no longer need to be transformed. 

\begin{algorithm}
\caption{Bird's Eye View of the Tiled QR Decomposition~\cite{buttari}}
\label{CHalgorithm}
\begin{algorithmic}[1]
\Statex {\bf Data}  Matrix $\+A \in \R^{M \times N}$ split into $R \times C$ blocks or tiles. 
\Statex {\bf Result} Matrix $\+A$ has been transformed into $\+R \in \R^{M \times N}$ from a QR factorization.
\vspace{.2in}
\For{ Blocks $\+A_{kk}$ along the diagonal}
\State Compute the QR Decomposition
\[ \+Q_{kk},\+A_{kk} \leftarrow \+Q_{kk}, \+R_{kk}= qr(\+A_{kk}) \] 
\hspace{.15in} updating $\+A_{kk}$ with the upper triangular factor $\+R_{kk}$.  
\For{Blocks $\+A_{kj}$ to the right of $\+A_{kk}$}
\State Use the projection matrix $\+Q_{kk}$ to update
\[ \+A_{kj} = \+Q_{kk}^T \+A_{kj} \]
\EndFor
\For{Blocks $\+A_{ik}$ below $\+A_{kk}$}
\State Compute the QR Decomposition with \quotes{coupling}
\[ \+Q_{ik}, \begin{bmatrix} \+A_{kk} \\ \+A_{ik} \end{bmatrix} \leftarrow \+Q_{ik}, \+R_{ik}   = qr\bigg(\begin{bmatrix} \+A_{kk} \\ \+A_{ik} \end{bmatrix}\bigg) \] 
\For{Blocks $\+A_{ij}$ to the right of $\+A_{ik}$}
\State Use the projection matrix $\+Q_{ik}$ to update
\[ \begin{bmatrix} \+A_{kj} \\ \+A_{ij} \end{bmatrix}  = \+Q_{kk}^T \begin{bmatrix} \+A_{kj} \\ \+A_{ij} \end{bmatrix}  \]
\EndFor

\EndFor
\EndFor

\end{algorithmic}
\end{algorithm}

\subsubsection{Tiled QR Algorithm applied to RPCA}
 Although the out-of-core QR algorithm generally works on two-dimensional tiles (i.e. row and column tiles), for our purposes, it suffices to take C=1, as we will see in Algorithm~\ref{RPCA-OQR}.     The reason for this is that the RPCA framework requires an in-core $\overline{K} \times P$ SVD at the end; thus, $\overline{K}$ inherently cannot be very large.  At the same time, we are considering the context of many samples (large $N$).  In other words, we assume that $\+Y \in \R^{N \times \overline{K}}$ will be a tall-and-skinny matrix.  As a result, we may simply formulate slices of the dataset $\+X$ such that the first slice has at least $\overline{K}$ rows (i.e., $M_{11} > \overline{K}$).    The setting where C=1 greatly simplifies the algorithm and reduces the number of matrix multiplication operations.   (Note, however, that for a slightly different problem, it would be similarly possible to set R=1 to handle matrices which were short-and-fat.)


%

\subsection{Introducing Large-Sample Randomized PCA (LS-RPCA): Randomized PCA with simultaneous out-of-core QR decomposition} 

As mentioned in Section~\ref{introducingLSRPCA}, if our randomly projected data $\+Y \in \R^{N \times \overline{K}}$ is still too large to fit into memory, then we need an out-of-core QR algorithm to implement the RPCA algorithm.  How should this be integrated into the standard implementation of RPCA, which was provided in Algorithm~\ref{RPCA}, in order to handle datasets with many samples?  

\begin{figure}
\centering
\begin{tikzpicture}
  \node[draw,fill=black,text=white]  (X) at (0,0) {X};
  \node at (1,.2) {$X\Omega$};
  \node[draw,fill=gray,text=white] (Y) at (2,0) {Y};
  \node at (3,.2) {$QR$};
  \node[draw] (R) at (4,0) {R};
  \node at (4.2,-.6) {$YR^{-1}$};
  \node[draw,fill=gray,text=white] (Q) at (3,-1) {Q};
  \node at (2.0,-.6) {$YR^{-1}$};
  \node at (2.5,-1.6) {$Q^T X$};
  \node at (1,-1) {$Q^T X$};
  \node[draw] (B) at (1,-2) {B};
    \node at (1.5,-2.7) {$SVD$};
  \node[draw] (V) at (1,-3.5) {V};
  \node[draw,fill=gray,text=white] (Xproj) at (-1,-5) {$X^{proj}$};
    \node at (-.75,-2) {$XV$};
       \node at (.5,-4.25) {$XV$};
        
  
  \draw[->,thick,draw=blue] (X) to (Y);
  \draw[->,thick,draw=blue] (Y) to (R);
  \draw[->,thick,draw=blue] (Y) to (Q);
  \draw[->,thick,draw=blue] (R) to (Q);
   \draw[->,thick,draw=blue] (Q) to (B);
   \draw[->,thick,draw=blue] (X) to (B);
  \draw[->,thick,draw=blue] (B) to (V);
  \draw[->,thick,draw=blue] (V) to (Xproj);
  \draw[->,thick,draw=blue] (X) to (Xproj);
 
\end{tikzpicture}
\caption{ \emph{Workflow for a standard RPCA implementation} Darker matrices are larger than lighter matrices.  (Black matrices are $N$ by $P$, grey matrices are $N$ by $\overline{K}$, and white matrices are $P$ by $\overline{K}$) }
\label{RPCAFlow}
\end{figure}
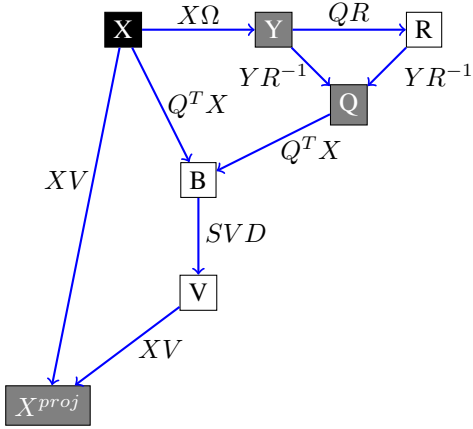

\begin{algorithm}
\caption{Large Sample RPCA (LS-RPCA)}
\label{RPCA-OQR}
\begin{algorithmic}[1]
\Statex {\bf Data}  Dataset $\+X \in \R^{N \times P}$ stored in $S$ horizontal slices 
\Statex \quad\quad \; Target dimensionality $K$ 
\Statex \quad \quad \; Oversampling dimensionality $\overline{K}$ 
\Statex {\bf Result} Projection matrix $\+V \in \R^{P \times K}$  
\vspace{.2in}
\State Form the random projection matrix 
\[ \+\Omega = randn(P,\overline{K}) \]
\State Use first data slice to initialize algorithm
\begin{align*}
\+Y_1&=\+X_1 \+\Omega\\
\tilde{\+Q},\+R&=qr(\+Y_1)\\
\+A&=\+Y_1^T\+X_1\\
\end{align*}
\For{ $s  \in \{2, 3, \hdots, S\}$  } 
\State Randomly project data slice $s$
\[ \+Y_s = \+X_s \+\Omega \] 
\State Update accumulator matrix for the Halko alg. 
\[ \+A \; \pluseq \+Y_s^T \+X_s \]
\State Update R for the out-of-core QR algorithm
\[ \tilde{\+Q},\+R = qr( \begin{bmatrix}  \+R \;\\ \+Y_s \end{bmatrix} )\] 
\EndFor
\State Obtain projection matrix through truncated SVD
\begin{align*}
 \+B &= (\+R^{-1})^T \+A \\
  \+U_{k,k} \+D_{k,k} \+V^T_{k,p} &= svd(\+B) \\
  \end{align*}
\end{algorithmic}
\end{algorithm}

We might initially consider a naive (modular) integration of the two algorithms, whereby we simply substitute the QR step of the RPCA algorithm with an out-of-core variant.  To see the problems with this approach, consider Figure~\ref{RPCAFlow}, which shows the workflow for RPCA.   Note that a direct, modular implementation of RPCA successively constructs necessary matrices from the earlier formed matrices: $\+Y$ from $\+X$, $\+Q$ from $\+Y$, and $\+B$ from $\+Q$ and $\+X$, and so forth.   With this in mind, a direct modular approach has two problems for datasets with many samples: (1) it would require two passes through the dataset X to get the SVD; and (2) various stages of the algorithm would require simultaneously storing two different very large datasets from the collection \{$\+X \in \R^{N \times P}, \+Y \in \R^{N \times \overline{K}}, \+Q \in \R^{N \times \overline{K}}$\}.   Although of course $\+Y$ and $\+Q$ are smaller than $\+X$, they still require large amounts of memory, given that there are many ($N$) rows.  In fact, $\+Y$ and $\+Q$ can easily consume more memory than $\+X$ if $\+X$ is sparse.   

 We solve this problem by creating a single-pass, low-storage algorithm for interfacing the Halko RPCA algorithm with the out-of-core, tiled QR algorithm.  The large-sample randomized PCA algorithm handles datasets with large samples by intelligently embedding an out-of-core QR decomposition into a  randomized PCA framework.     The key insights are two-fold:
 \begin{enumerate} 
 \item By backwards substitution through the steps of the standard RPCA implementation given in Algorithm~\ref{RPCA}, and performing linear algebra operations of transposing and inverting, we can express the matrix $\+B$ as the following sum over horizontal \quotes{slices} of the data matrix $\+X$:
 \begin{equation}
 \label{B}
 \+B = (\+R^{-1})^T \ds\sum_{s=1}^{\text{nSlices}}  \+\Omega^T \+X_s^T \+X_s  
 \end{equation}
where we can operate upon one slice of the data $\+X_s$ at a time, and where $\+R \in \R^{\overline{K} \times \overline{K}}$ is both invertible and small. 
 \item  As we construct $\+Y_s :=   \+X_s \+\Omega $ in Equation~\ref{B}, we can simultaneously compute the necessary information for the out-of-core QR decomposition in Algorithm~\ref{CHalgorithm}.   
 \end{enumerate}

The full LS-RPCA algorithm is provided in Algorithm~\ref{RPCA-OQR}.    Overall, the algorithm extends RPCA to datasets with large sample sizes, which would break a standard implementation of RPCA.   The algorithm integrates RPCA and out-of-core QR into a single-pass algorithm, and formulates the crucial (small) $\+B \in \R^{\overline{K} \times P}$  matrix accumulatively and directly from the original dataset, so there is no need to write large temporary dense by-products (such as $\+Y$ and $\+Q$) to disk.

\section{Experiment 1: Large Malware Dataset}

In this experiment, we address Question \#1 on a real-life cybersecurity dataset which is sufficiently large to call for application of the LS-RPCA algorithm developed in Section~\ref{introducingLSRPCA}. 

\subsection{Data and Method}  \label{e1_datamethods}


Our dataset is N=11,725,193 heterogeneous portable executable files obtained ``in the wild" and determined to be either malicious or clean.   Each portable executable file was represented as 98,450 features previously demonstrated to be relevant to whether a file was malicious or not.  The features were mixed continuous and binary, and therefore were represented as (Float32,Int64) Sparse CSC matrices.  The density level across the dataset was approximately .0224, so the sparse matrix included over 25 billion nonzero values.  The dataset comprised approximately 88 GB of data in memory.  All computations were performed on an Amazon EC2 r3.4 instance with 16 cores and 122 GB of RAM.   As discussed in the introduction, the dataset was sufficiently large to require out-of-core treatment to perform expensive linear algebra computations such as QR decompositions; hence we applied our LS-RPCA algorithm (Algorithm~\ref{RPCA-OQR}) to determine the approximate principal components.  



To evaluate the quality of the different projection techniques for supervised learning, we adhered to the following pipeline: (1) \emph{normalization}, where, to preserve sparsity, we normalized only the non-zero values of the continuous features by subtracting off the mean and dividing by 2 standard deviations; (2) \emph{projection}, where we projected with either random projections or the large-sample RPCA method of Algorithm~\ref{RPCA-OQR} (using a Gaussian RP for the random projection step in both techniques), (3) \emph{renormalization}, due to the fact that the projections undo the original normalizations by creating linear combinations of the original features; (4) \emph{training of classifier}, where we trained a logistic regression classifier on a training set which was composed by randomly selecting, without replacement, 80\% of available samples; (5) \emph{test set evaluation}, where we apply the trained classifier to the remaining 20\% of samples.  

\subsubsection{Oversampling of the column space}
For the RPCA projection, Halko et al.~\cite{halko_big} recommend power iterations (see Section~\ref{RPCA_section}); that is, pre-multiplication of the dataset in Step 2 of Algorithm~\ref{RPCA} such that $\+Y=(\+X\+X^T)^q\+\Omega$, after $q$ iterates.   However, for  truly large datasets (such as those considered here), performing such power iterations  can be prohibitive.   Thus, in lieu of power iterations, we attempted ``oversampling" in the data matrix range approximation step (Step 2 of Algorithm~\ref{RPCA}).    By oversampling, we mean that we are setting $\overline{K}$ to a larger value than $K$.  Although as a general rule of thumb, it is not necessary to set $\overline{K}$ much higher than $K+5$ or $K+10$~\cite{halko_pca}, ultimately, the necessary amount of oversampling depends on the size of the matrix, the decay properties of the spectrum, and the choice of the random projection matrix~\cite{mahoney}.   Thus, we compare three variants of LS-RPCA:
\begin{enumerate}
\item \emph{LS-RPCA} conservatively sets $\overline{K}=5000$; for most values of K, this creates a great deal of oversampling that may compensate for slow spectral decay. 
\item \emph{LS-RPCA\_double} sets $\overline{K}=2K$; this is the procedure specificially recommended by~\cite{halko_big} for truncated SVD's.
\item \emph{LS-RPCA\_minimal} sets $\overline{K}=K$; this minimizes computational effort. 
\end{enumerate} 

\subsection{Results and Discussion}

\begin{figure}
\centering
\includegraphics[height=3.6in]{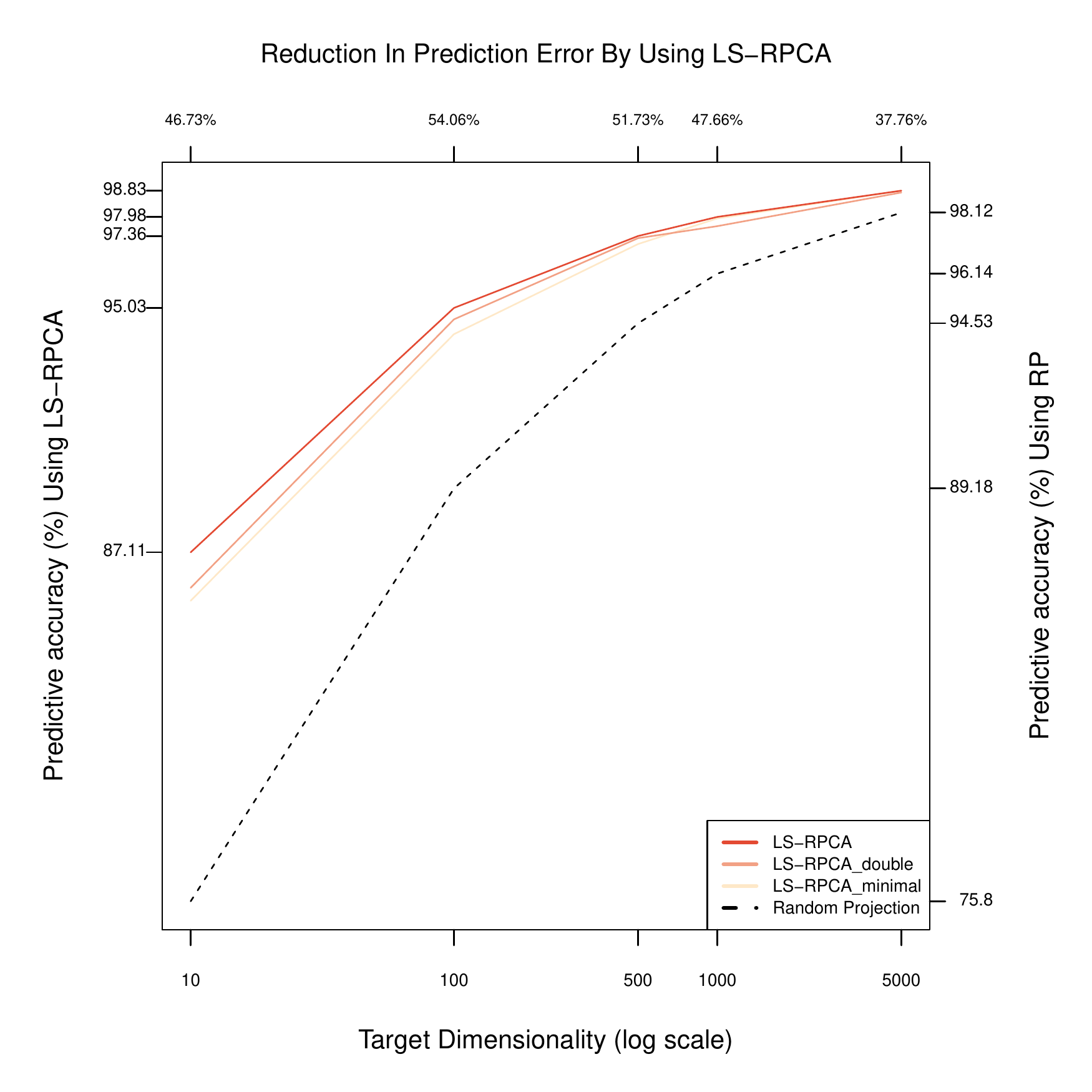}
\caption{ \emph{Randomized PCA produces better projections than random projections.} The plot shows the predictive performance of the classifier, applied to the malware dataset, across a range of target dimensionalities, after projecting with either RPCA or RP.    Due to the large size of our data set, in particular the large multiplicative factor between the number of samples and the number of target dimensions, we applied our LS-RPCA algorithm (Algorithm~\ref{RPCA-OQR}) to determine, approximately, the top $K$ principal components. The reduction in prediction error at each target dimensionality is shown above the plot. }  
\label{RPvsRPCA}
\end{figure}

Does randomized PCA provide a higher quality dimensionality reduction than random projections? In Figure~\ref{RPvsRPCA}, we show the results of our experiment, comparing the predictive performance of our dataset in a classifier, across a range of target dimensionalities, after projecting with either RPCA or RP.    
We first note that, as predicted by~\cite{halko_big}, the impact of heavily oversampling appears to be relatively minor, although oversampling does produce better results overall. 

More importantly, we see that, across the full range of target dimensions investigated, RPCA indeed produces higher quality dimensionality reduction than random projections.  The downstream benefit of using RPCA for dimensionality reduction, rather than random projections, is to reduce prediction error by anywhere between 36\% and 54\% (depending on the value of $K$).   This is an enormous difference.    Although it does appear that the quality of random projections is converging to the quality of RPCA as $K$ increases, the values of $K$ studied cover reasonable values that researchers use for real large-scale problems (e.g. the Dahl et al.~\cite{dahl} malware classification paradigm reduced dimensionality to $K=4000$ before feeding the features into a neural network).  

\begin{figure}
\centering
\includegraphics[height=3.6in]{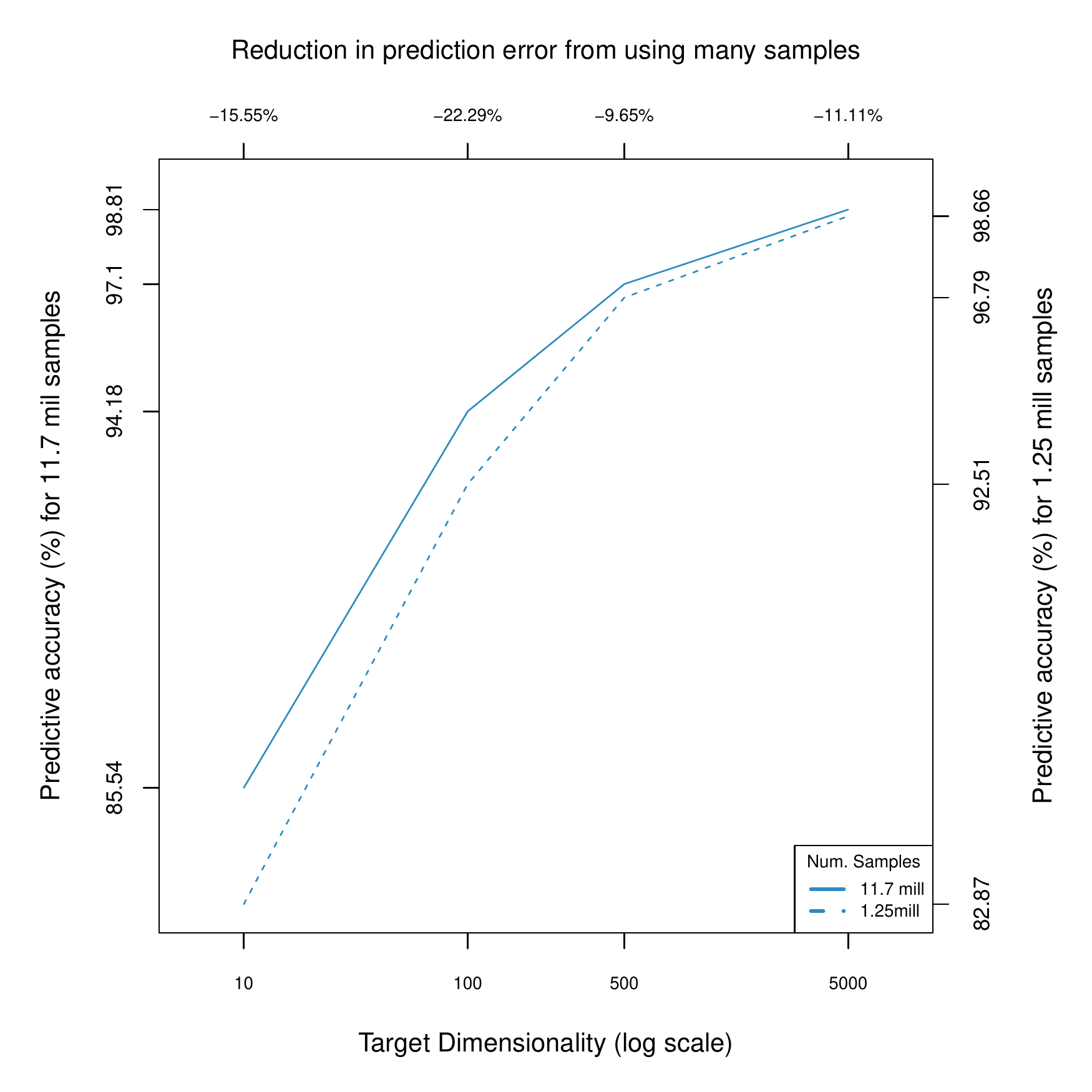}
\caption{\emph{The LS-RPCA algorithm, by expanding the number of samples that can be processed, improves the quality of a RPCA-based projection.}  The plot shows predictive performance after training with the full set of available samples, which requires our LS-RPCA procedure (Algorithm~\ref{RPCA-OQR}), vs. training with the approximately largest subset that could be processed by Baseline RPCA (Algorithm~\ref{RPCA}).The reduction in prediction error at each target dimensionality is shown above the plot. } 

\label{largesample}
\end{figure}

Thus, the current study certainly suggests that impactful studies (e.g., the Dahl et. al paradigm, and others like it) could potentially improve classification accuracy by utilizing RPCA instead of RP.  We note that performing RPCA on all 2.6 million samples in~\cite{dahl}, in our computing context, would require LS-RPCA.  Moreover, we can show that the ability of LS-RPCA to handle many samples actually improved downstream predictive performance in the current experiment. In Figure~\ref{largesample}, we compare classifier performance when processing the full 11.7 million samples, rather than restricting to a 1.25 million sub-corpus (approximately the maximum number of samples that could be processed by Algorithm 1 to trace out these curves). We found that the ability to process the full corpus reduces prediction error by roughly 10\%-20\%. By construction, LS-RPCA incurs no additional computational complexity.

\section{Experiment 2: Kaggle Competition} \label{e2_datamethods}

One might wonder whether some idiosyncratic properties of the proprietary industrial dataset from Experiment 1 could have been responsible for the relative effectiveness of RPCA.  Thus, here we replicate the findings  on a second, publicly available dataset, while testing the robustness of the effect against two different classifiers and multiple possible data representations (sparse vs. dense; binary vs. continuous). 


\subsection{Data and Method}

This dataset is the 2015 Kaggle Microsoft Malware Classification Challenge dataset.   The goal is to classify a sample into one of 9 possible malware families (\emph{ramnit, lollipop}, etc.).    The training set is N=10,868 malware samples.  Following one common strategy among top-performing teams in the competition, we construct predictive features by transforming the hexadecimal representations of each sample's binary content (with the PE header stripped out) into 4-grams.  Moreover, to motivate dimensionality reduction, we extract out a relatively large number of (the 100,000 most common) 4-grams. 


We create four versions of the dataset.   For the \emph{binary features} ({\tt BINARY}) dataset,  $\+X_{(i,j)}$ is set to 1 if the $i$th sample contains the $j$th 4-gram, and to 0 otherwise. For the \emph{count features} ({\tt COUNT})  dataset, $\+X_{(i,j)}$ is set to the number of times that the $j$th 4-gram appeared in the $i$th sample.   The resulting matrices have density 0.0915.   We then normalize the datasets, so that the right singular vectors approximated by Algorithm 1 will equal the principal components. The massive size of the dataset in Experiment 1 led us to perform \emph{sparse normalization} ({\tt SPARSE}), where we normalized only the non-zero values from each column of continuous features (and thus left binary features unchanged).     The smaller size of the Kaggle dataset means that, in this experiment, we may also apply \emph{dense normalization} ({\tt DENSE}) , where each column $\+x_j$ is normalized in the standard way, by subtracting $mean(\+x_j)$ and dividing by $2 \cdot sd(\+x_j)$, therefore densifying a sparse dataset.  For our purposes, these four representations allow us to address two questions: \emph{Does the advantage of RPCA hold for both sparse and dense data matrices?} and \emph{Does it hold for both continuous and binary datasets?}

We project the data with either ({\tt RP}) or ({\tt LS\_RPCA}), using a Gaussian RP for the random projection step in both techniques.   Based on Experiment 1, we implement minimal oversampling; i.e.  $\overline{K}=K$.   We then apply two classifiers:  \emph{multi-class logistic regression} (paralleling Experiment 1) and  \emph{xgboost} (as used by the Kaggle winners).  Because the purpose of this experiment was to compare dimensionality reduction techniques, we simply fixed Xgboost parameters to those used by the team \emph{Marios \& Gert}~\cite{2nd}, and logistic regression parameters to the default values from Python's scikit-learn implementation for a multi-class scenario (using the limited-memory Broyden-Fletcher-Goldfarb-Shanno optimization algorithm).   We evaluate model performance with the \emph{multi-class logarithmic loss} function used by Kaggle, and report the average loss across 5-fold cross validation.

\subsection{Results and Discussion}

\begin{figure}
\centering
\includegraphics[height=3.6in]{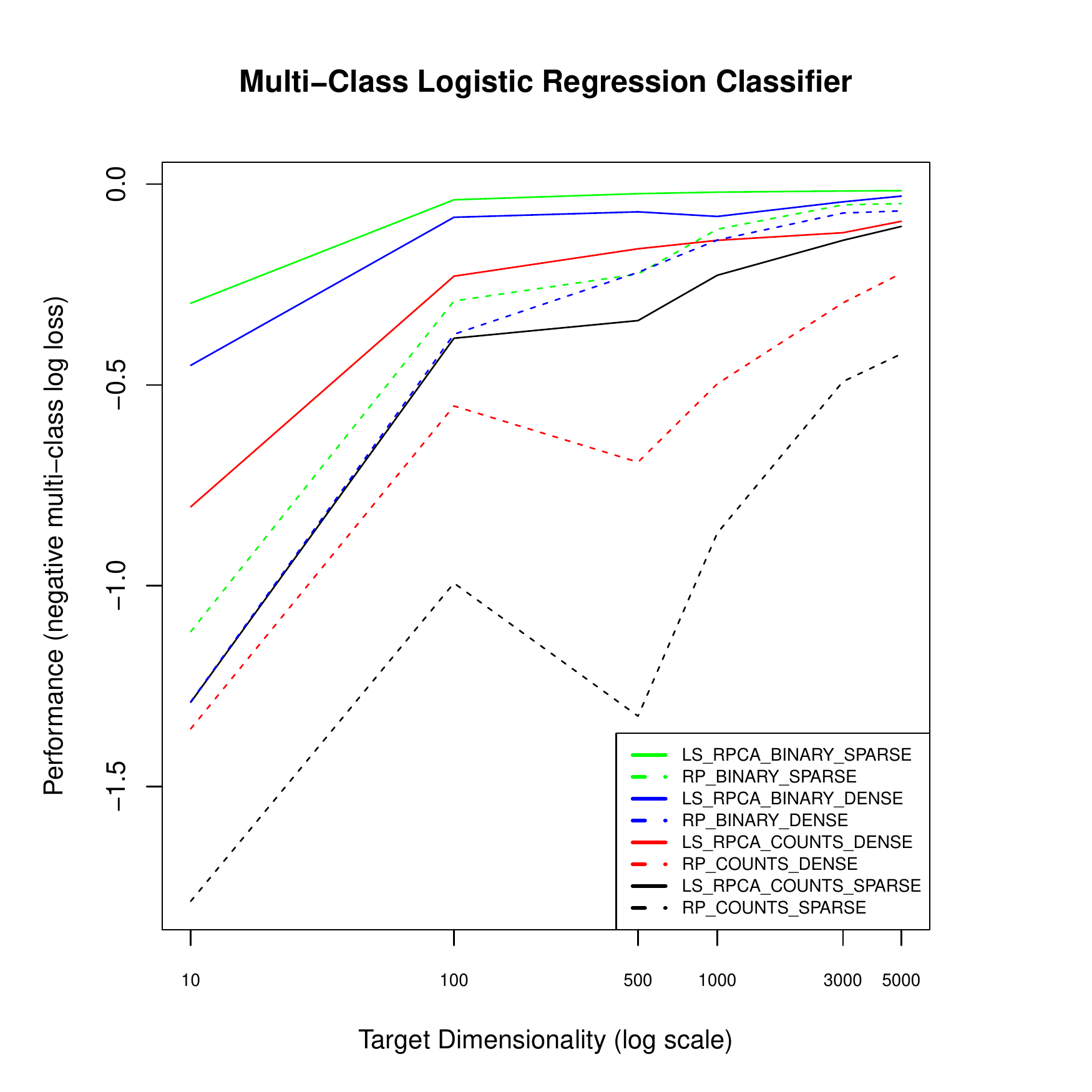}
\caption{\emph{RPCA provides a higher-quality dimensionality reduction than RP for training a multi-class logistic regression classifier on the Kaggle malware classification task.} }
\label{kaggle_lr}
\end{figure}

\begin{figure}
\centering
\includegraphics[height=3.6in]{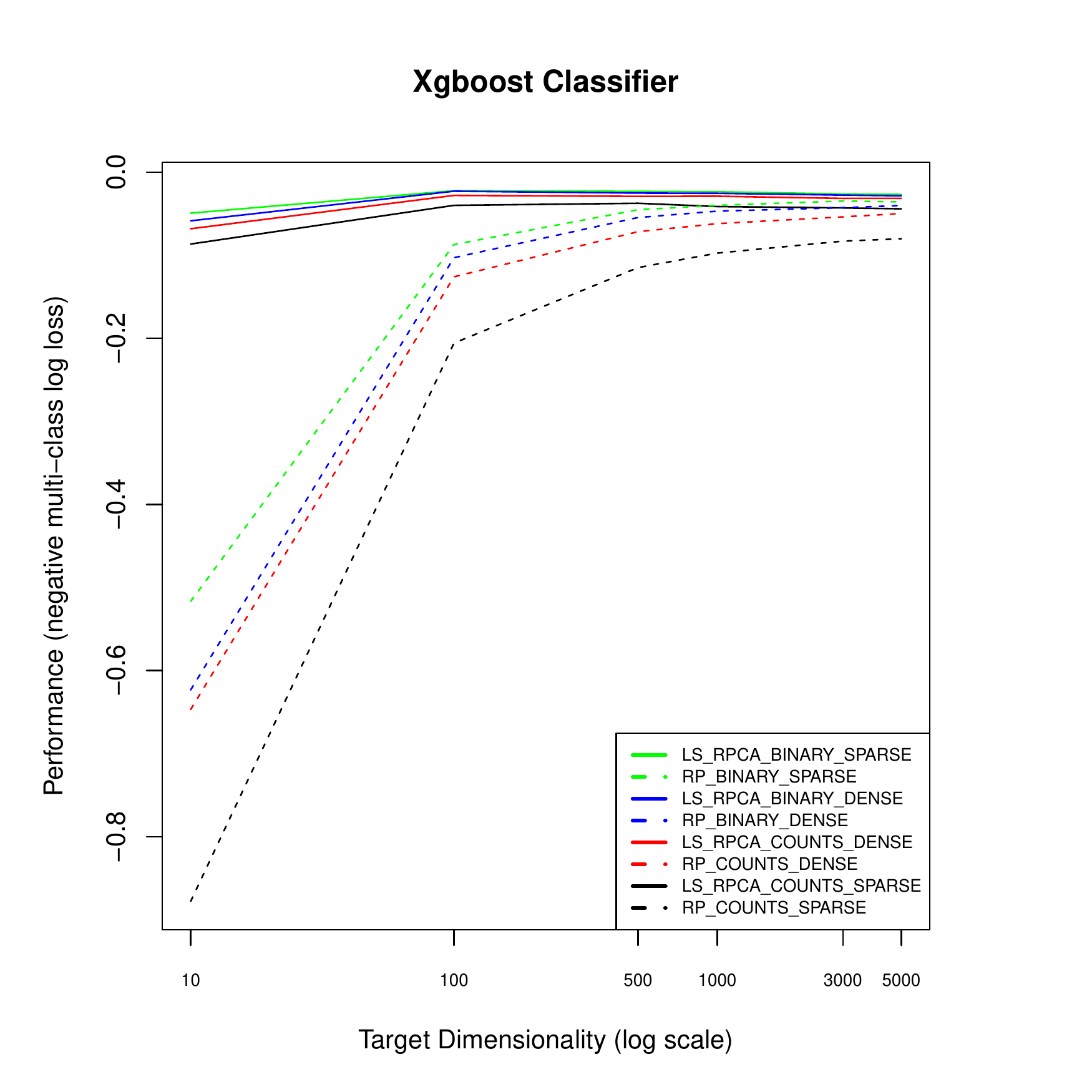}
\caption{\emph{RPCA provides a higher-quality dimensionality reduction than RP for training an xgboost classifier on the Kaggle malware classification task.} }
\label{kaggle_xg}
\end{figure}

In Figures~\ref{kaggle_lr} and ~\ref{kaggle_xg}, we show the performance of multi-class logistic regression and  xgboost classifiers, respectively, when these classifiers are applied to the Kaggle malware family prediction problem after using RPCA or RP as dimensionality reduction techniques.    The plots show the mean cross-validation error (expressed as negative multi-class log loss, so that higher values mean better performance) across a range of target dimensionalities.  In Figure~\ref{kaggle_lr}, note that for every colored pair of curves representing a fixed feature representation strategy, dimensionality reduction using RPCA produces better classification performance than dimensionality reduction using RP.   In Figure~\ref{kaggle_xg}, note that the primary discriminant across performance curves is the dimensionality reduction strategy. 

\section{Conclusion}

In this paper, we compared the performance of RPCA and RP in terms of reducing dimensionality of a large dataset for downstream classification.   We also developed a variant of the RPCA algorithm, called large sample RPCA (LS-RPCA), which allows randomized PCA to be applied to datasets with many samples.  Using LS-RPCA, we discovered that we could create better dimensionality reductions than with RP across multiple datasets, feature representations, and classifiers. The relationship of \emph{random} RPCA to RP appears to mirror the relationship of \emph{classical} PCA to RP~\cite{fradkin}; the advantage holds when the target dimensionality $K$ is low or moderate, but disappears as $K$ grows. 


We mention two future directions: (1) Is it possible to scalably integrate sparse PCA into the LS-RPCA framework?  Sparse principal components would create a projection matrix that is smaller in size and easier to store. However, most sparse PCA frameworks, even when they scale to large data sets, don't scale well to large $K$; e.g. the method of~\cite{takac} requires deflation methods and the method of~\cite{hastie_spca} would require iteratively running $K$ blocks of elastic nets.    (2) Is it possible to gracefully integrate a parallel SVD into the algorithm?  As it currently stands, RPCA algorithms face a bottleneck in that an in-core SVD must be performed on a $P \times \overline{K}$ matrix.   This limits the range of possible dimensionality reduction sizes for datasets with large number of predictors;  e.g. the current method applied to a dataset with 3 million predictors at the current sparsity level analyzed on an Amazon EC2 r3.4 instance would only be able to reduce the feature dimensionality to approximately 100-200 predictors.

%
%

\end{document}